# Integrating Spatiotemporal Vision Transformer into Digital Twins for High-Resolution Heat Stress Forecasting in Campus Environments

Wenjing Gong[a], Xinyue Ye[a,*], Keshu Wu[a], Suphanut Jamonnak[a], Wenyu Zhang[a], Yifan Yang[b], Xiao Huang[c]

[a] *Department of Landscape Architecture and Urban Planning & Center for Geospatial Sciences, Applications and Technology, Texas A&M University, College Station, USA*
[b] *Department of Geography, Texas A&M University, College Station, USA*
[c] *Department of Environmental Sciences, Emory University, Atlanta, USA*

**Abstract:** Extreme heat events, exacerbated by climate change, pose significant challenges to urban resilience and planning. This study introduces a climate-responsive digital twin framework integrating the Spatiotemporal Vision Transformer (ST-ViT) model to enhance heat stress forecasting and decision-making. Using a Texas campus as a testbed, we synthesized high-resolution physical model simulations with spatial and meteorological data to develop fine-scale human thermal predictions. The ST-ViT-powered digital twin enables efficient, data-driven insights for planners and stakeholders, supporting targeted heat mitigation strategies and advancing climate-adaptive urban design. This campus-scale demonstration offers a foundation for future applications across broader and more diverse urban contexts.

## 1. Introduction

Extreme heat events, intensified by climate change, pose growing health and socio-economic risks, particularly in urban and campus environments (Anderson and Bell 2011; Chu and Rotta Loria 2024; Georgescu, Broadbent, and Krayenhoff 2024). Unlike other hazards, heat is invisible, exacerbating its impact on vulnerable populations and contributing to rising emergency visits and health issues such as cardiovascular and mental health disorders (Campbell et al. 2018; Mallen et al. 2020). The complexity of urban heat dynamics necessitates proactive, data-driven planning to enhance resilience and mitigate risks.

Campuses pose distinct heat exposure challenges due to their spatial structure, activity patterns, and population dynamics. Diverse microclimates, shaped by open spaces, greenery, and outdoor activity areas, interact with frequent movement between buildings, exposing students and staff to fluctuating thermal conditions (Jiang et al. 2024; Mallen et al. 2020). Compared to urban cores, campus shading and cooling infrastructure is often insufficient for extreme heat events (Göçer et al. 2019). At the same time, modeling thermal comfort at the micro scale remains challenging due to the fine-grained spatial and temporal variability of key variables such as air temperature and wind speed. Prior studies have employed both physics-based models and data-driven AI approaches to assess urban heat stress at high resolution (Bruse and Fleer 1998; Jia et al. 2024). However, the complex and dynamic microclimatic conditions on campuses call for more adaptive, scalable, and context-specific solutions.

Digital twins offer a promising solution for mitigating extreme heat exposure on campuses by providing virtual replicas that simulate and manage the built environment (White et al. 2021; Ye et al. 2023). These data-driven platforms integrate real-time inputs, models, and simulations to assess environmental stressors and inform decision-making (Omrany and Al-Obaidi 2024; Riaz, McAfee, and Gharbia 2023; Ye et al. 2024). By enabling dynamic analysis of heat mitigation strategies, digital twins allow stakeholders to predict thermal conditions, evaluate interventions, and enhance climate resilience without physical implementation.

Despite their potential, climate digital twins face two key challenges. First, traditional physical climate models are computationally intensive and costly, limiting their feasibility for real-time predictions and integration into digital twins (Hu et al. 2023; Xiaojiang Li et al. 2024). Their high resource demands and slow simulation speeds hinder rapid decision-making and reduce accessibility, particularly in resource-constrained settings. Second, many existing digital twins primarily serve as static visualization tools, lacking predictive capabilities to simulate heatwave progression or the impact of mitigation strategies (P. Liu et al. 2023; Ramani et al. 2023). Without forward-looking modeling, these systems fail to support proactive decision-making, leaving stakeholders reliant on representations of past and present conditions rather than anticipating future risks.

Artificial intelligence (AI) models offer a promising solution to the limitations of traditional climate modeling by reducing computational demands while maintaining accuracy, enabling real-time, data-driven heat predictions within digital twins (Bibri et al. 2024; Ketzler et al. 2020; Lehtola et al. 2022). However, two key challenges remain. First, AI models relying solely on meteorological data often fail to capture the nonlinear and nonstationary dynamics of atmospheric processes, leading to potential prediction errors (Jing Shi, Guo, and Zheng 2012; Zhu et al. 2024). Second, effective climate prediction requires integrating diverse climate and spatial datasets to model spatiotemporal dependencies critical for understanding localized heat variations (Zhu et al. 2024). Addressing these challenges is essential to enhancing AI-driven high-resolution climate forecasts and advancing their role in digital twin applications for climate resilience.

To overcome these challenges, this study presents a human-centric climate digital twin framework, using the Texas A&M University (TAMU) campus as a testbed. High-resolution 3D campus models, derived from LiDAR point clouds and land cover maps, alongside meteorological data, were integrated to represent urban geometry and localized climatic contexts. The Spatiotemporal Vision Transformer (ST-ViT) model incorporated multimodal data, including Universal Thermal Climate Index (UTCI) (Bröde et al. 2012) mappings simulated via physics-based methods, to capture fine-scale spatiotemporal dependencies and enable real-time, high-accuracy heat predictions, with “real-time” denoting rapid processing and inference within seconds (Jihao Shi et al. 2025; Fujiwara et al. 2024). By combining physics-based simulations to retain atmospheric dynamics with a Transformer architecture that models long-range dependencies across space and time, the ST-ViT model enhances climate prediction precision and efficiency. Integrating the ST-ViT model into a digital twin platform, this study establishes a data-driven decision-support tool for heat resilience planning. The research makes three key contributions:

- We proposed a human-centric climate digital twin framework that integrated the ST-ViT model with real-time prediction capabilities and practical components to support informed decision-making and planning for TAMU campus heat resilience.
- We developed an advanced ST-ViT model that coupled output from physics-based simulations to preserve atmospheric dynamical physical characteristics and leveraged self-attention mechanisms to provide a computationally efficient approach for precise predictions of human heat stress.
- We leveraged multimodal data to identify high-resolution campus hot spots across different stages of a heatwave and diurnal cycles, providing detailed insights to guide resource allocation, planning, and strategies for mitigating extreme heat impacts.

## 2. Literature review

### 2.1. Physical, AI, and hybrid models in urban climate prediction

In the field of urban climate modeling, two main approaches are currently predominant: physics-based climate models and data-driven AI models. The former includes established models such as ENVI-met (Bruse and Fleer 1998), SOLWEIG (Lindberg, Holmer, and Thorsson 2008), WRF (Weather Research and Forecasting) (Skamarock et al. 2019), and Fluent (Fluent 2011). These models rely on complex mathematical calculations and numerical simulations, offering high spatial resolution and detailed meteorological predictions, while often incurring high computational costs and time demands (Han et al. 2024; L. Zheng and Lu 2024). With the advancement of AI, an increasing number of researchers have turned to machine learning and deep learning models for urban climate predictions (Fujiwara et al. 2024; Jia et al. 2024). These AI models leverage their ability to capture nonlinear relationships between inputs and outputs while reducing computational resources and runtime.

Among AI architectures for spatiotemporal climate prediction, various approaches have been explored. Convolutional Neural Networks combined with Long Short-Term Memory networks (CNN-LSTM) are commonly used to extract spatial and temporal features but often struggle with capturing long-range temporal dependencies (Xinyu Li et al. 2023). U-Net variants with temporal extensions, such as ConvLSTM or 3D convolutions, can preserve fine spatial resolution, but their encoder-decoder structure and repeated processing of large feature maps often lead to high computational and memory costs (Briegel et al. 2023). Recently, Vision Transformers, particularly ST-ViT, have emerged as promising alternatives. Their self-attention mechanisms enable efficient parallel modeling of both local and global spatiotemporal dependencies, allowing flexible focus across time and space (Lin et al. 2023). Unlike convolution-based models with fixed receptive fields, ST-ViT can dynamically learn contextual relationships, making it well-suited for complex urban microclimate patterns that exhibit heterogeneous spatial structures and temporal lags. However, relying solely on spatial or meteorological data may cause AI models to fail in retaining the physical characteristics of atmospheric dynamics embedded in nonlinear and nonstationary climate data, potentially resulting in significant prediction errors (Jing Shi, Guo, and Zheng 2012).

Hybrid models, which leverage external coupling by combining the outputs of one or more base models as features and integrating them with advanced models, have recently gained attention (Wu, Wang, and Zeng 2022; Zhu et al. 2024; Briegel et al. 2024). This approach allows for the integration of strengths from multiple individual models, creating complementary advantages and optimizing the overall model structure. This potential offers a new perspective and methodology for achieving accurate and efficient urban climate predictions, as well as for building robust digital twin systems. For instance, Briegel et al. utilized SOLWEIG (SOlar and LongWave Environmental Irradiance Geometry) model input and output data as the dataset for a U-Net model, achieving high-precision predictions of mean radiant temperature ($T_{mrt}$) at a 1-meter scale (Briegel et al. 2023). However, this study did not account for the temporal dependencies in the data; temporal information was linearly transformed and then added to the compressed latent spatial representations. Another study introduced a framework that used building energy modeling and CFD (Computational Fluid Dynamics) simulation results as datasets for a graph attention network to predict outdoor thermal comfort at an urban micro-scale (L. Zheng and Lu 2024). While this model focused on capturing the spatial interrelationships of urban features, it lacked considerations for the cooling effects of greenery and did not incorporate temporal dependencies, limiting its ability to predict thermal comfort across different time steps. Recently, Zhu et al. proposed a hybrid model combining WRF and temporal fusion transformers (TFT), where WRF outputs served as the prediction dataset for the TFT model (Zhu et al. 2024). This model achieved high predictive performance in forecasting urban air temperatures in central Guangzhou, China. However, the resolution was limited to 0.5 km, and air temperature alone does not fully represent human thermal comfort.

These studies collectively underscore the necessity and potential of developing computationally efficient and accurate frameworks to support urban climate modeling. Compared to traditional physics-based models, data-driven approaches provide a scalable and flexible foundation for building responsive, high-resolution digital twins in urban climate contexts. When coupled with outputs from physics-based models to form hybrid models, they gain additional advantages. However, data-driven methods in urban climate modeling are still in the early stages of development (Yang et al. 2023), and research on spatiotemporal predictions of human thermal comfort at fine scales, particularly their integration into digital twins, remains limited.

## 2.2. Digital twins in urban climate and heat mitigation

The digital twin has emerged as a transformative tool, integrating digital innovations with urban operations to support planning and climate analysis (Peldon et al. 2024; Xia et al. 2022).

Studies highlight their effectiveness in visualizing thermal environments, forecasting climate conditions, and assessing intervention strategies (Cárdenas-León et al. 2024; T. Liu and Fan 2023). However, many digital twins primarily focus on real-world data visualization, integrating 3D city models, thermal imaging, and meteorological data to map microclimates and temperature variations (Ramani et al. 2023). While valuable for monitoring, their full potential for proactive heat mitigation planning remains unrealized due to the lack of predictive and decision-support capabilities.

Prediction and simulation are fundamental to digital twins, enabling dynamic climate forecasting and mitigation planning (Deren, Wenbo, and Zhenfeng 2021). For instance, a smart city digital twin framework incorporating the SARIMA time series model and crowd simulation demonstrated how digital twins can help urban officials identify, predict, and mitigate extreme heat exposure (Pan et al. 2024). However, statistical models often fail to capture complex nonlinear relationships and struggle to leverage spatial characteristics inherent in climate data. In urban climate modeling, AI-driven spatiotemporal predictions offer a promising approach to dynamically evaluate and forecast human thermal comfort within digital twins (Čulić et al. 2021; Karyono et al. 2024). For example, Liu et al. applied a GraphSAGE model with street-view imagery to predict outdoor thermal comfort, suggesting its integration into urban digital twins (P. Liu et al. 2023). While this method effectively captures spatial dependencies, it overlooks temporal dynamics, which are essential for modeling seasonal, diurnal, and transient climate patterns necessary for accurate heat predictions.

Moreover, there is a notable lack of hybrid models that integrate physics-based simulations with AI-driven predictions within digital twin platforms. This gap limits the ability to preserve atmospheric dynamics and leverage the complementary strengths of both approaches, particularly in complex urban environments where high-resolution climate predictions are critical. University campuses, such as TAMU, represent thermal vulnerability hotspots, with dense populations, extensive outdoor activities, and diverse building configurations creating localized heat exposure patterns. Conventional models often fail to resolve these patterns at a fine scale, leading to gaps in accurately assessing campus heat risks. Additionally, many existing studies do not fully integrate their models into digital twin platforms or translate insights into actionable heat mitigation strategies. This is particularly consequential for campus environments, where real-time decision support for facility managers is essential to mitigating heat stress impacts. These challenges highlight the urgent need for AI-powered campus digital twins to bridge the gap between advanced modeling techniques and practical applications in urban climate resilience.

## 3. Data and methodology

### 3.1. Study area and analytical framework

The TAMU main campus in College Station, Texas, USA, spans about 6.13 km² and serves as the study area (Supplementary Fig. S1). With a subtropical climate of hot summers and mild winters, the region is projected to experience intensifying extreme heat, surpassing early 20th-century levels by 2036 (Texas 2036 2023). The campus's varied land cover and building layouts provide an ideal setting for analyzing extreme heat impacts on human heat stress and advancing digital twin development for informed decision-making.

This study presents a novel human-centric climate digital twin framework for urban heat stress monitoring and prediction, integrating data collection, physical simulation, spatiotemporal analysis, deep learning models, and digital twin application into a cohesive system (Fig. 1). It begins with the collection of spatial and meteorological data to establish the foundational data layer. Subsequently, through a physically-based human heat stress simulation module, we utilized the SOLWEIG model to calculate $T_{mrt}$ and generated UTCI maps. The third module analyzes the spatiotemporal variation of heat stress, visualizing the dynamic changes in the thermal environment. The system further employs an ST-ViT model, combined with UTCI output from the physics-based approach, efficiently capturing complex spatiotemporal dependencies in multimodal data to deliver rapid thermal predictions. Finally, the framework leads to a digital twin application that transforms theoretical analysis into practical heat exposure response strategies, facilitating campus heat exposure overview, route planning, and advanced response predictions for more resilient urban thermal management solutions.

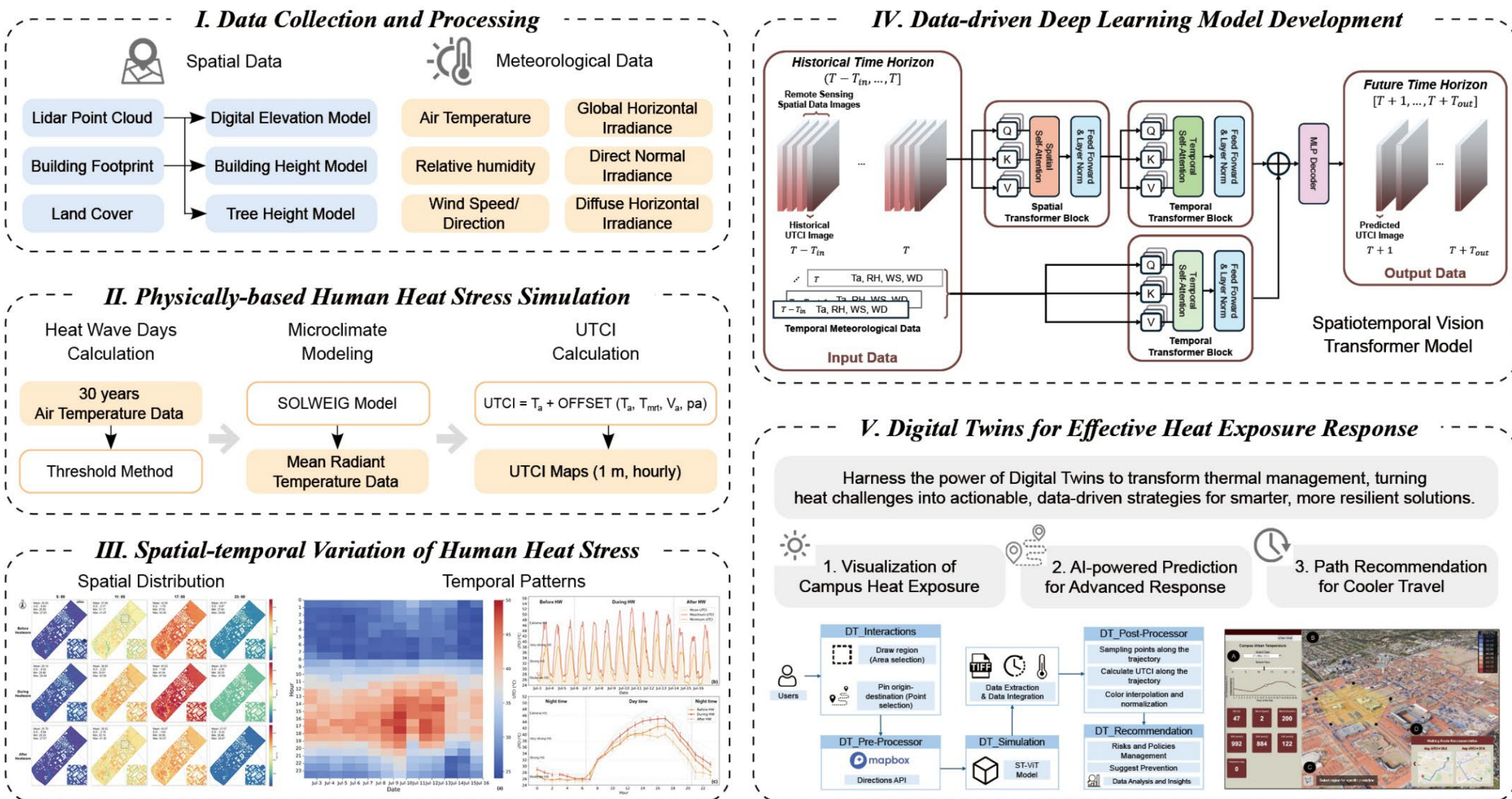

Fig. 1. Analytical framework.

### 3.2. Heat wave days calculation

To explore human outdoor heat stress under extreme conditions, the heat wave days were calculated to define the simulation period. They were identified based on 30 years (1991–2020) of air temperature data from a campus meteorological station ("ASOS" n.d.). The 98th percentile of daily maximum temperatures (38.33°C) served as the heat wave threshold (Z. Zheng, Zhao, and Oleson 2021). Periods in 2022 exceeding this threshold for at least three consecutive days were analyzed, with the longest event, July 6–13, lasting eight days, defined as the heat wave (Meehl and Tebaldi 2004). The simulation period included three days before and after the heat wave, spanning 14 days to assess heat exposure impacts.

### 3.3. Physically-based human heat stress simulation

3.3.1. Physical simulation method of human heat stress

This study utilized the UTCI, a widely recognized index for quantifying human heat stress and assessing urban microclimate impacts on outdoor thermal comfort (Yang et al. 2023). UTCI calculations mainly rely on $T_{mrt}$, air temperature, relative humidity, and wind speed (Bröde et al. 2012), with $T_{mrt}$ being the most critical input (Xiaojiang Li et al. 2024). We employed the physics-based SOLWEIG model (Lindberg, Holmer, and Thorsson 2008) to simulate hourly $T_{mrt}$ at a 1-meter resolution across a 14-day heat wave period, covering day and night. The SOLWEIG model has been widely applied and validated globally for urban microclimate simulations (Hu et al. 2023; Chaowen and Fricker 2021). While SOLWEIG supports UTCI calculations only at specific points, a custom script was developed to generate UTCI maps for the entire study area. The SOLWEIG model configuration and justification for high-resolution UTCI simulation are detailed in Supplementary Note S1.

3.3.2. Data for human heat stress measurement

This study utilized spatial and meteorological data to measure human heat stress. Spatial data included LiDAR point clouds, Digital Elevation Model (DEM), building footprints, and land cover data, all essential inputs for the SOLWEIG model. LiDAR and DEM data (1-meter resolution) were obtained from USGS ("USGS" n.d.), with the LiDAR processed using PointCNN ("Point Cloud Classification Using PointCNN" n.d.) to classify tree points and generate a tree height model. Building footprints from TAMU were combined with DEM and LiDAR data to create a calibrated building height model, while land cover data was also sourced from TAMU. Meteorological data included hourly air temperature, relative humidity, wind speed, and wind direction from an ASOS station on campus ("ASOS" n.d.), as well as hourly radiation data from NREL's NSRDB ("NSRDB" n.d.) for the heat wave periods, including global horizontal irradiance (GHI), direct normal irradiance (DNI), and diffuse horizontal irradiance (DHI).

### 3.4. ST-ViT model for human heat stress forecast

3.4.1. Data and processing

This study developed the ST-ViT model to enable fine-scale spatiotemporal UTCI prediction within digital twin frameworks. A detailed summary of data characteristics and roles is provided in Supplementary Table S1. It includes four spatial datasets and seven temporal meteorological variables, which are also required inputs for the SOLWEIG model. To better capture the high-resolution spatiotemporal dynamics of human heat stress, historical UTCI map sequences were also included as input features. While SOLWEIG simulates each UTCI map independently, using the sequence allows the model to learn temporal patterns, such as diurnal heat retention, that enhance predictive accuracy beyond what static spatial features and spatially uniform meteorological inputs alone can offer. Notably, other variables required for SOLWEIG simulations, such as the Sky View Factor (SVF), derived from building height and tree height through additional pre-processing, were excluded to reduce redundancy and computational cost.

Spatial and temporal data were normalized for training, with predicted UTCI maps denormalized for interpretability. Using 336 hourly time steps, the model predicted UTCI for the next 24 hours ($T_{\mathrm{out}}$ = 24) based on the preceding 24 hours ($T_{\mathrm{in}}$ = 24). Different sliding window steps were tested to evaluate the model's sensitivity to input configurations. The dataset was split chronologically (first 70% for training), with data augmentation (random cropping of 64×64 subregions) applied to improve generalization and robustness (Dyk and Meng 2001; Maharana, Mondal, and Nemade 2022). Testing involved dividing the study area into overlapping 64×64 subsets with 10-pixel overlaps, and merging predictions to calculate final evaluation metrics.

### 3.4.2. Model architecture

The ST-ViT model is a transformer-based, decoder-only architecture designed for efficient and precise spatiotemporal dependency modeling (N. Liu et al. 2021; Shim et al. 2023). Inputs, including spatial images and temporal meteorological features, were embedded into a shared hidden space through linear projection layers. The model employed three parallel attention mechanisms: spatial attention for local and global spatial relationships, and two temporal attention mechanisms for capturing long-term dependencies of human heat stress and meteorological features. After parallel processing, spatial-temporal features and temporal meteorological features were fused through an additive mechanism. Temporal features were expanded to match the spatial resolution, and the unified representation was passed through a final linear projection layer, ensuring the preservation of spatial and temporal resolutions. Further details on the model architecture are provided in Supplementary Fig. S2.

### 3.4.3. Model implementations

We compared the ST-ViT model with two benchmark models, CNN+LSTM and U-Net+LSTM, commonly used for spatiotemporal prediction (Yin et al. 2023; Xinyu Li et al. 2023). All models were implemented in Python with PyTorch and trained on two NVIDIA A800 Tensor Core GPUs (80GB memory each). Mean Squared Error Loss (MSELoss) was used as the loss function with the Adam optimizer (learning rate=0.0001) for optimization (Kingma and Ba 2017). Early stopping was applied to prevent overfitting by monitoring validation loss. Model performance was evaluated using Root Mean Squared Error (RMSE), Mean Absolute Error (MAE), and Mean Absolute Percentage Error (MAPE). The calculation formula is shown from Eq. (1) to Eq. (4). Additional details of the models can be found in Supplementary Table S2.

$$MSE = \frac{1}{m}\sum_{i=1}^{m}(\hat{UTCI}_i - UTCI_i)^2 \tag{1}$$

$$RMSE = \sqrt{\frac{1}{m}\sum_{i=1}^{m}(\hat{UTCI}_i - UTCI_i)^2} \tag{2}$$

$$MAE = \frac{1}{m}\sum_{i=1}^{m}\left|\hat{UTCI}_i - UTCI_i\right| \tag{3}$$

$$MAPE = \frac{100\%}{m}\sum_{i=1}^{m}\left|\frac{\hat{UTCI}_i - UTCI_i}{UTCI_i}\right| \tag{4}$$

where $m$ is the total number of pixels; $\hat{UTCI}_i$ denotes the ST-ViT predicted UTCI for pixel $i$; and $UTCI_i$ denotes simulated UTCI based on SOLWEIG for pixel $i$.

### 3.5. Application: Interactive campus climate digital twin

#### 3.5.1. Overview of platform architecture

The digital twin platform integrates high-resolution UTCI predictions from the ST-ViT model into an interactive web-based interface, transforming complex thermal data into actionable insights for campus heat mitigation. Through photorealistic 3D Tiles technology, the platform creates a detailed virtual representation of the campus environment, which is dynamically overlaid with UTCI predictions visualized as color heatmaps at a spatial resolution of 1 meter and hourly temporal granularity. This interactive visualization enables users to quickly identify critical heat exposure zones across the campus environment. Beyond basic visualization, the platform offers two advanced decision-support functions: Path Recommendation—generating optimized routes to minimize heat exposure, and Predictive Heatmap—utilizing the ST-ViT model to forecast future thermal conditions. This seamless integration of thermal modeling with intuitive visualization and practical features creates an effective tool for making informed thermal comfort choices in the campus environment.

#### 3.5.2. Technologies and frameworks used

The interactive web-based digital twin platform integrates advanced technologies and frameworks to ensure scalability, usability, and accuracy in data processing, modeling, visualization, and user interaction. The backend architecture consists of two main components. First, a Python-based data preprocessing pipeline handles the computation of UTCI predictions using the ST-ViT model on tile-based imagery. This pipeline leverages specialized libraries like NumPy for numerical computations and GeoPandas for spatial data operations, enabling precise overlay of UTCI predictions onto campus geographical layers. Second, a Flask framework serves as a lightweight backend server, hosting the prediction endpoints and facilitating seamless integration between the ST-ViT model and the web interface. The frontend architecture was built on React JS for responsive user interactions across devices, utilizing Deck.GL (a powerful JavaScript library) to render photorealistic 3D tiles and dynamically overlay UTCI predictions as interactive heat maps onto the campus model.

3.5.3. Platform deployment

The interactive web-based campus digital twin platform is deployed on Amazon Web Services (AWS), leveraging scalable cloud infrastructure. AWS EC2 instances provide the core computational resources for prediction and visualization tasks. The backend services are containerized using Docker to ensure consistent performance across different environments and enable efficient resource management and automated scaling. This cloud-based deployment architecture not only ensures current platform reliability but also supports future scalability through the seamless integration of additional data sources and features.

## 4. Results

This section began with a spatiotemporal analysis of human heat stress simulated using physical climate models. A prediction model database was then constructed based on these results and input into the ST-ViT model for UTCI forecasting. The ST-ViT model's performance was validated against physical model results, achieving accurate and rapid predictions. Finally, the model was integrated into the climate digital twin platform to support risk warnings and planning for human heat stress.

### 4.1. Spatiotemporal variation of human heat stress based on SOLWEIG

Based on the temporal and diurnal UTCI patterns observed in Supplementary Figs. S3 and S4, which show that 17:00 is typically the peak UTCI hour across the heatwave, we selected four representative times (05:00, 11:00, 17:00, and 23:00) to show the spatial distributions of UTCI in Fig. 2. During the heatwave stage, UTCI at 17:00 reached up to 47.99 °C, compared to 45.90 °C before the heatwave and 43.07 °C after it, indicating intensified thermal stress during the peak period. Notably, spatial heterogeneity was highest at 11:00 across all heatwave stages, with the largest standard deviation observed during the heatwave period (2.22), compared to before (2.17) and after (2.15) the heatwave. Impervious surfaces exhibited consistently higher UTCI, while vegetated, water, and shaded zones remained relatively cooler across all stages and times.

The distribution of human heat stress categories across the campus based on pixel proportion is illustrated in Supplementary Fig. S5 (see Supplementary Table S3 for category definitions). During all heatwave periods, no campus areas experienced "no thermal stress" in the daytime, indicating widespread discomfort. Before the heatwave, much of the campus faced strong or very strong heat stress. During the heatwave, extreme heat stress became prevalent, while nighttime stress levels also increased. Post-heat wave, thermal conditions improved, with no extreme heat stress in the daytime and a rise in moderate heat stress and no thermal stress at night, indicating partial recovery.

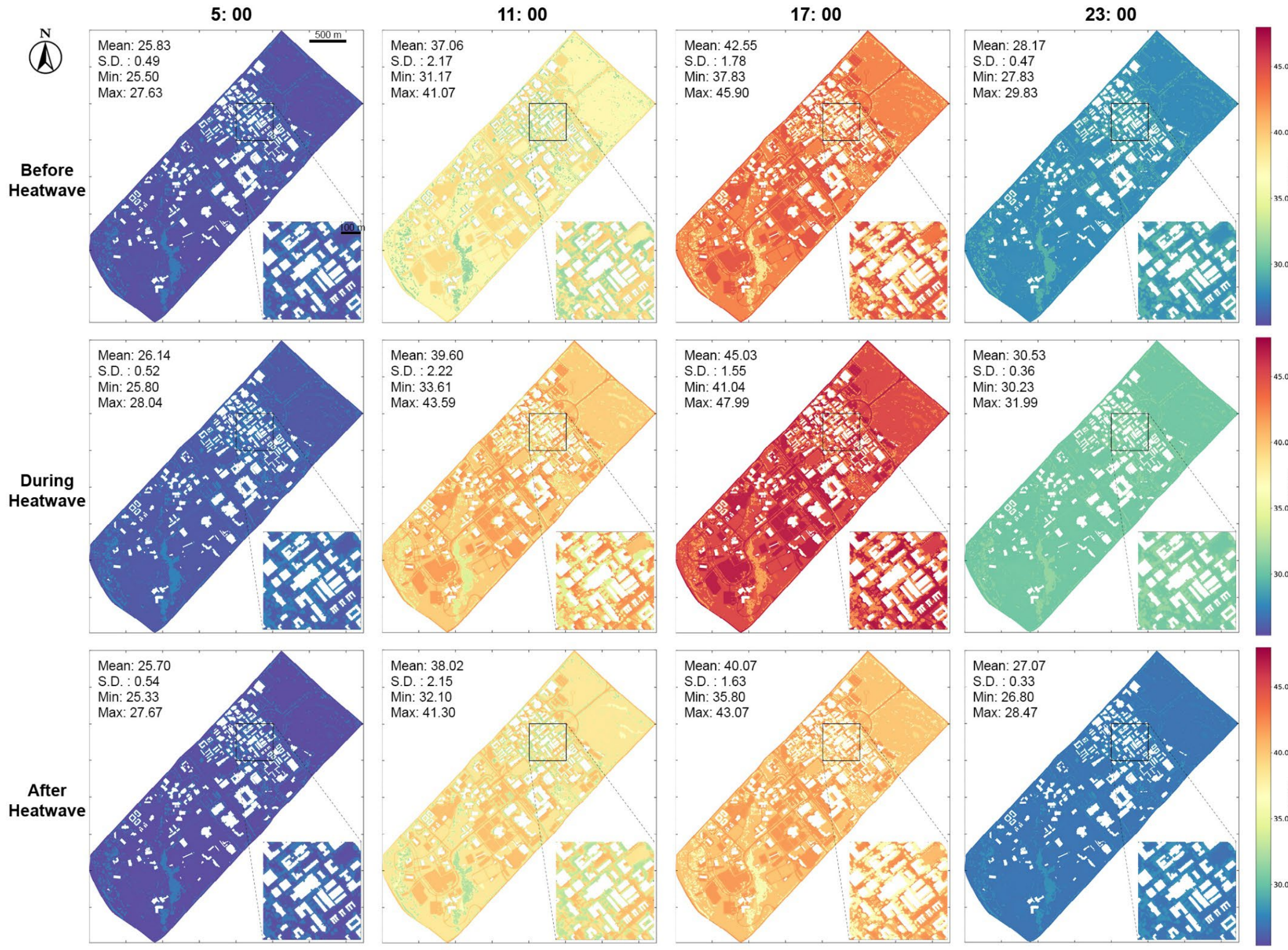

Fig. 2. UTCI distributions at four representative hours during different heatwave stages, with each map showing the average UTCI across all days within the stage at the specified hour.

## 4.2. ST-ViT model performance

### 4.2.1. Model comparisons

We evaluated the performance of deep learning models across various time step lengths and selected the one with the highest predictive accuracy for further analysis (results for the ST-ViT model are provided in Supplementary Table S4). Supplementary Table S5 compares the overall performance, where the ST-ViT model outperformed the baseline models (CNN+LSTM and U-Net+LSTM) across all metrics, achieving the lowest RMSE (2.163°C), MAE (1.770°C), and MAPE (5.811%). These results demonstrate the ST-ViT model's superiority in UTCI prediction. Additionally, the ST-ViT model exhibited real-time capability, predicting a 1200×1200 pixel image in about 7.2 seconds—16 times faster than the QGIS UMEP SOLWEIG-based approach, and 680 times faster when SOLWEIG preprocessing is considered (e.g., SVF calculation) on the same computer.

#### 4.2.2. Spatiotemporal prediction effectiveness of the ST-ViT model

Fig. 3(a-p) illustrates the diurnal distribution of UTCI across typical campus areas on July 15th at four intervals, comparing ST-ViT predictions (a-d) with SOLWEIG simulations (e-h) and their differences (i-p). Both methods effectively captured the influence of urban morphology on UTCI, including building and tree shadows, as well as thermal comfort variations across different land cover types. Nighttime predictions (05:00 and 23:00) showed strong agreement between the two models. However, the ST-ViT model underestimated UTCI at 23:00, while it overestimated UTCI at 05:00, 11:00, and 17:00 in most areas, particularly in areas with dense vegetation and building shadows. Additionally, the ST-ViT model tended to smooth daytime UTCI predictions, particularly for extreme values, resulting in narrower distribution ranges and more concentrated peak values.

Fig. 3(q-r) quantifies ST-ViT prediction errors across land cover types, revealing significant temporal dependence and distinct diurnal cycles. The MAE was generally higher during the daytime, peaking around noon and lowest during the early morning (2:00-6:00). Among land cover types, tree-covered areas exhibited the largest errors and most significant fluctuations, with median MAE around 2.0 and peak values of approximately 3.8 during 12:00-14:00. In contrast, water surfaces consistently showed the smallest errors throughout the diurnal cycle, indicating better model performance in predicting water body thermal environments. Paved surfaces and grass followed the average MAE pattern throughout the day, aligning with the overall temporal trends.

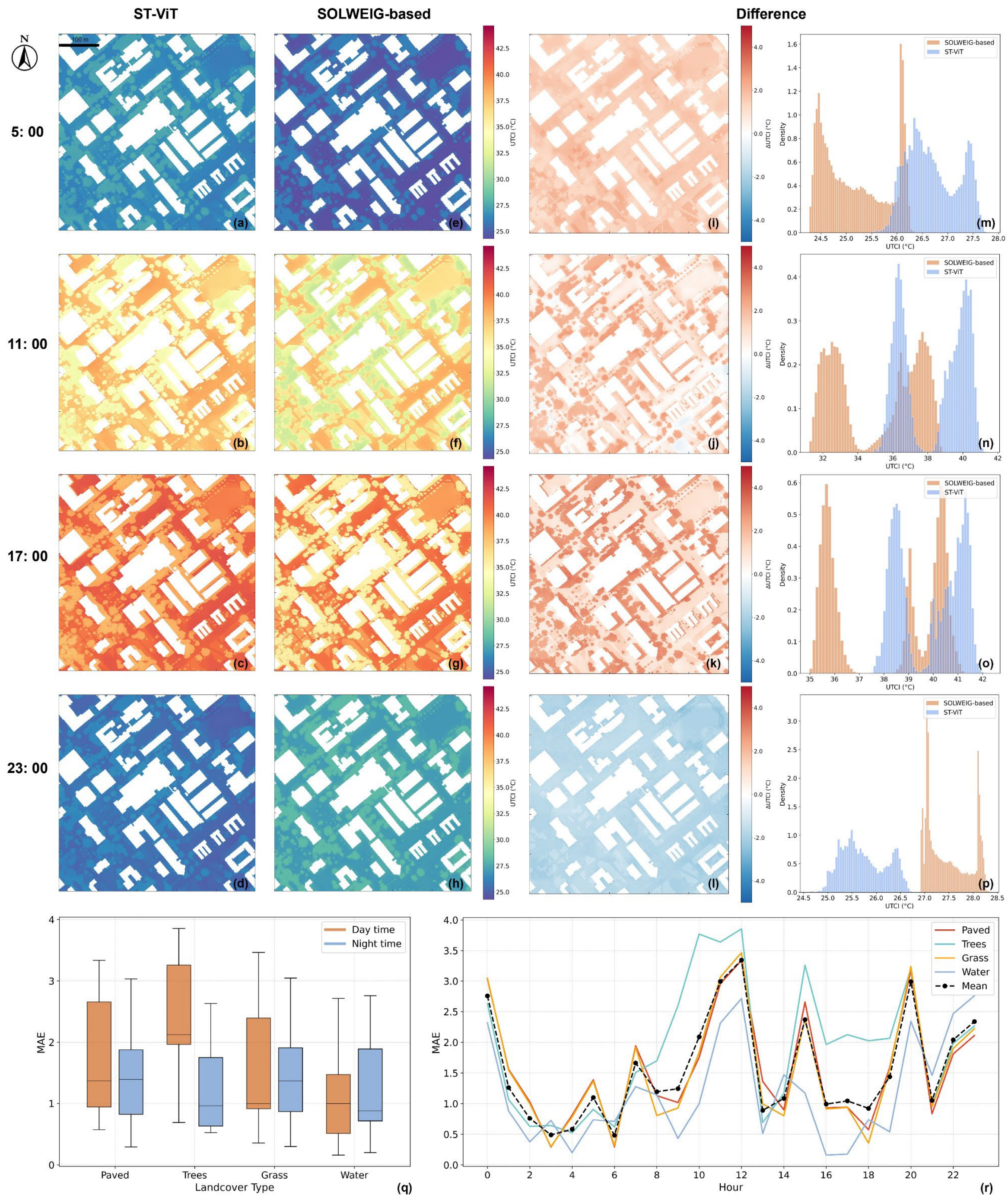

Fig. 3. Diurnal distribution of UTCI predictions from ST-ViT (a-d), SOLWEIG-based simulations (e-h), and their differences (i-p) across typical campus areas (500 × 500 m) on July 15th, 2022; UTCI prediction errors (MAE) of ST-ViT model (q) during daytime and nighttime, and (r) hourly variation across land cover types for the entire study area. See Supplementary Fig. S6 for the landcover classifications.

### 4.3. Dynamic digital twin platform for effective response

#### 4.3.1. User interface

The dynamic digital twin platform provides an interactive web-based interface that integrates comprehensive analytical and operational capabilities to address urban thermal comfort challenges. The user interface features four core modules. First, the Time of Day Controller and Data Summary View (Fig. 4A) enables users to adjust the temporal granularity (e.g., hourly or daily) for analyzing dynamic UTCI changes, helping understand thermal comfort patterns for resilient planning. Second, the Photorealistic 3D Tiles Map (Fig. 4B) visualizes the spatial distribution of UTCI at a high resolution of 1 meter, using color-coded heatmaps to display thermal hotspots. Third, the Region Selection Function (Fig. 4C) allows users to select specific areas for ST-ViT-powered UTCI simulations, enabling tailored analyses for specific user-defined scenarios. Lastly, the Path Recommendation Module (Fig. 4D) generates optimized walking routes with average UTCI values based on user-defined origin and destination. Together, these modules create a highly interactive interface that supports decision-making for urban thermal comfort management.

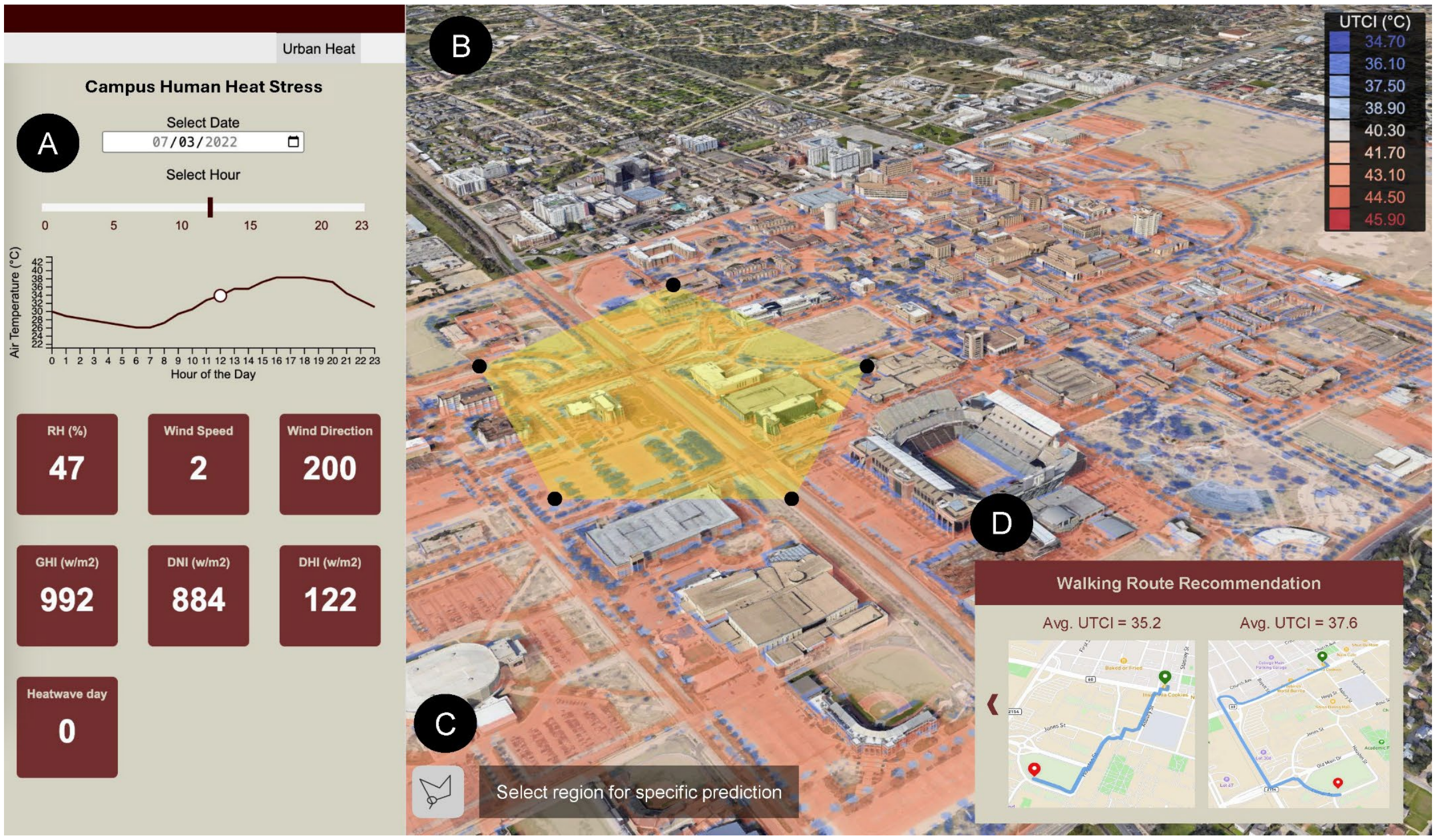

Fig. 4. Digital twin platform interface: (A) Time of Day Controller and data summary view: allows users to adjust hour granularity along with detailed data summary; (B) Photorealistic 3D Tiles map: visualize UTCI in color heatmap per 1-meter; (C) Function: Select the region for UTCI forecasting powered by ST-ViT model; and (D) Function: path recommendation based on user's OD selection.

4.3.2. Key technological components

The back-end operation flow of the platform, illustrated in Fig. 5, relies on three core technological components: high-resolution visualization, integrated prediction models, and path recommendation algorithms. For visualization, the platform leverages 3D tiled maps combined with multi-layer data integration to dynamically display thermal comfort distributions. Heatmaps, using color coding at a 1-meter resolution, allow users to detect spatial disparities in thermal comfort, while the interface supports interactive functions such as panning, zooming, and overlaying additional layers. These features enable users to identify problem areas and assess the potential impact of various mitigation strategies effectively.

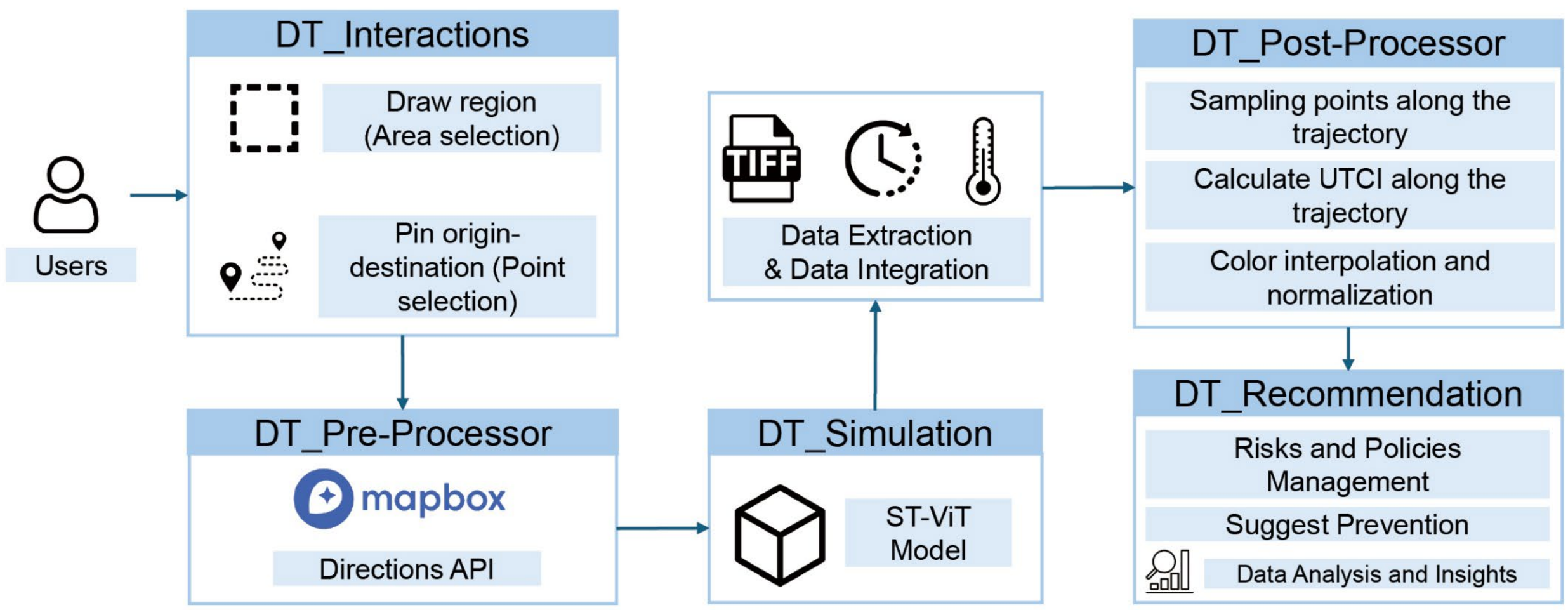

Fig. 5. Back end of digital twin platform.

The second component is the integration of advanced prediction models. The platform incorporates pre-trained ST-ViT to predict and simulate dynamic UTCI conditions. The workflow for model integration includes multi-source data preprocessing, model inference, and post-processing for visual output. Input data, such as weather parameters, urban morphology, and vegetation indices, are harmonized in terms of spatial and temporal resolution before being fed into the ST-ViT model. The predictions are then processed into user-friendly visual outputs, such as heat maps and simulation results. This seamless integration of deep learning models enhances computational efficiency while enabling spatiotemporal forecasting of thermal conditions up to 24 hours in advance, equipping users with predictive insights to better plan for and mitigate heat stress impacts in dynamic campus environments.

Path recommendation is another key feature of the platform, leveraging grid-based algorithms to optimize walking routes for improved thermal comfort. The grid structure is defined as $G = (V, E)$, where $V$ represents a set of nodes (grid cells) and $E$ represents a set of edges (connections between adjacent grid cells). A walking trajectory $P$ is defined as an ordered node sequence traversing the grid: $P = (v_1, v_2, \ldots, v_n)$, where $v_1$ is the origin ($V_i$) and $v_n$ is the destination ($V_j$). The average UTCI for path $P$ is calculated by aggregating the node-specific thermal comfort weights along the trajectory. The formula is:

$$\text{Avg UTCI}(P) = \frac{1}{|P|}\sum_{k=1}^{|P|} W(v_k) \quad (5)$$

where $|P| = n$: Total number of nodes in path $P$; $W(v_k)$: UTCI value of the node $v_k \in V$.

The platform leverages algorithms such as Dijkstra, A*, and multi-criteria optimization to generate alternative routes. These algorithms consider both the shortest distance and thermal comfort along the route, allowing users to adjust the weight of each criterion to suit their preferences. This feature provides users with actionable insights for choosing convenient and comfortable walking paths.

## 5. Discussion

### 5.1. Campus heat stress patterns and mitigation implications

The spatiotemporal variations of UTCI, calculated based on SOLWEIG simulations, reveal that most campus areas experience strong human heat stress or higher risk levels, underscoring the urgent need for mitigation under extreme heat conditions. The findings emphasize the critical role of land cover types in shaping campus thermal conditions. Paved areas, especially large impervious surfaces like parking lots, consistently exhibit the highest levels of heat stress during the day. This observation aligns with previous studies highlighting the thermal burden of such surfaces, supporting the use of mitigation strategies—such as reflective coatings, permeable pavements, or shading structures—that have been shown in the literature to reduce surface temperatures and enhance thermal comfort (H. Li et al. 2013; Cheela et al. 2021).

Tree-covered areas and water bodies provide substantial daytime cooling benefits by reducing UTCI. However, during nighttime, dense tree canopies may inhibit radiative cooling and airflow (Wujeska-Klause and Pfautsch 2020), resulting in slightly elevated UTCI values. The potential side effects of nighttime warming can be mitigated through thoughtful landscape design, including the selection of tree species with good canopy ventilation, careful control of planting density, and the integration of open spaces (Norton et al. 2015).

### 5.2. Evaluating ST-ViT model performance: efficiency gains and systematic biases

In this study, we developed the ST-ViT model, a novel deep-learning approach for rapidly forecasting spatiotemporal heat stress patterns—unlike static map provision, which offers only snapshots in time. To the best of our knowledge, this is the first application of a Transformer-based architecture for micro-scale modeling of human heat stress at a 1-meter resolution on an hourly basis.

Integrated into a decision-support digital twin platform, the ST-ViT model enhances practical usability by effectively modeling UTCI and closely approximating the complexities of the SOLWEIG model, even with unseen data. The hybrid approach, leveraging physics-based SOLWEIG outputs as training data within the ST-ViT architecture, not only preserves essential atmospheric dynamics but also delivers substantial improvements in computational efficiency. This finding aligns with prior studies highlighting the advantages of hybrid models in balancing physical realism and operational feasibility (Zhu et al. 2024; Jihao Shi et al. 2025). By addressing key computational barriers that have traditionally hindered the deployment of fine-scale climate models, this method makes advanced thermal comfort analysis more accessible to campus planners and facility managers who may lack high-performance computing infrastructure. Furthermore, the ST-ViT model outperformed conventional architectures such as CNN+LSTM and U-Net+LSTM across all evaluation metrics, demonstrating the superior capacity of Transformer-based models to capture complex spatiotemporal dependencies in urban microclimate data.

Despite its advantages, the ST-ViT model exhibited certain systematic biases. It slightly overestimated UTCI during daytime hours (11:00 and 17:00) and early morning (05:00) across most areas, while underestimating it at 23:00. This temporal pattern suggests a degree of diurnal bias, similar to patterns reported in previous studies showing greater UTCI prediction errors during daytime hours in Germany (Briegel et al. 2024). This bias may reflect the model's incomplete representation of nighttime radiative cooling and thermal inertia processes. In addition, the model tended to overestimate UTCI values in areas characterized by dense vegetation and building shadows during daytime hours. This may indicate limitations in its ability to capture localized cooling effects and an incomplete representation of shading and emissivity in the input features (Briegel et al. 2023).

Moreover, the ST-ViT model was prone to smoothing extreme UTCI values, particularly during periods of peak thermal stress. This behavior likely reflects the model's predisposition to prioritize global patterns over localized microclimatic extremes. Similar tendencies have been reported in other deep learning models, including U-Net, which has shown systematic biases when predicting thermal extremes (Briegel et al. 2023).

### 5.3. Human-centered AI-based digital twins for planning and education

The AI-powered digital twin platform developed in this study represents an emerging paradigm for advancing climate-informed curricula and fostering climate-adaptive, sustainable campus environments. This approach also aligns with broader objectives of equity and technological democratization. Universities, as microcosms of urban systems, provide ideal testbeds for applying and teaching intelligent resilience technologies (Rosak-Szyrocka and Wolniak 2025). The ST-ViT-powered digital twin enables deeper exploration of the interconnections between climate, infrastructure, and human behavior, while fostering interdisciplinary collaboration across urban planning, computer science, and climate science. Furthermore, by lowering computational barriers, such platforms hold the potential to empower resource-constrained institutions to adopt similar capabilities through technology transfer and collaborative frameworks. This opens pathways for participation in advanced, participatory resilience planning and contributes to greater equity and accessibility across the higher education sector.

The human-centric design of the digital twin framework further enhances its educational and operational value across multiple dimensions. First, the selection of UTCI as the thermal comfort indicator, rather than basic air temperature metrics, better captures the nuances of actual human thermal perception (Yang et al. 2023). Its high spatial resolution (1 meter) enables person-scale insights into microclimatic variability. Second, the platform features an intuitive interface, spatially explicit thermal visualizations, and actionable functions that enable campus stakeholders to forecast dynamic thermal conditions, identify high-risk areas in advance, and explore alternative walking routes. These capabilities translate complex climate model outputs into accessible, practical insights, empowering users to proactively respond to extreme heat. The platform also lays the foundation for future extensions that could support scenario-based evaluations of heat mitigation strategies, reflecting a shift in climate adaptation from retrospective assessment to forward-looking, human-centered planning.

Finally, the deployment of AI-driven digital twins offers a transferable model for broader urban planning applications. Lessons learned from campus environments—whose spatial complexity, pedestrian dynamics, and diverse land uses closely mirror those of urban cores—can inform city-wide climate adaptation strategies, infrastructure resilience planning, data integration approaches, and the development of AI-driven digital twin systems at the city scale. In this way, campus digital twins contribute not only to localized climate risk management but also to the global knowledge base for sustainable urban futures amid escalating climate uncertainty.

### 5.4. Limitations and outlook

There are some limitations to this study. First, this study focuses on evaluating the performance of the ST-ViT model as a potential alternative to physics-based models, rather than reassessing the SOLWEIG model, which has already been extensively validated in various urban environments, as mentioned in Section 3.3. Second, improvements in model performance can be achieved through hyperparameter optimizations and architectural enhancements, as well as the inclusion of more detailed spatial and meteorological predictors (Zhu et al. 2024). However, the overall results also depend on the performance of the underlying SOLWEIG model.

Third, while the ST-ViT model shows strong performance within the TAMU campus, its generalizability to larger or more heterogeneous regions remains uncertain. Although random cropping of spatial subregions was used during training to improve local spatial generalization, the model has not been explicitly evaluated on entirely unseen spatial areas. Applying the model to regions with different urban morphology, climatic conditions, or data characteristics will require additional validation and potential adaptation to ensure transferability.

Furthermore, the ST-ViT model follows a sequence-to-sequence paradigm, using the preceding 24 hours of data to model spatiotemporal dependencies and forecast the subsequent 24 hours of UTCI. While the current implementation focuses on short-term forecasting due to data and computational constraints, the framework is designed to be scalable. It can be retrained on new datasets to accommodate different time periods and adapted to incorporate additional input variables through configuration adjustments. As more historical training data become available or stakeholder needs evolve, the model can also be extended for longer-term forecasting, with reduced reliance on recent inputs by leveraging learned seasonal and meteorological patterns to lower computational demands.

Finally, our current framework falls primarily within the predictive stage of the digital twin continuum. Future development could incorporate stakeholder-driven scenario simulation modules, intelligent agents, and true bidirectional connectivity, gradually advancing the system toward the simulative and cognitive stages to support more adaptive, evidence-based decision-making.

## 6. Conclusion

Extreme heat poses significant risks to public health and urban environments, particularly in complex and vulnerable settings like university campuses. Addressing these challenges requires innovative approaches that support sustainable campus planning and urban climate adaptation. In this study, we proposed an innovative climate digital twin framework that coupled physics-based simulations to construct a comprehensive database and developed the ST-ViT model for fine-scale spatiotemporal prediction of human heat stress. By incorporating the ST-ViT model into the digital twin framework, this research bridged the gap between advanced AI modeling and practical urban climate management. It demonstrated the potential of combining high-resolution predictive models with decision-support systems to equip campus stakeholders with effective tools to address the growing challenges of extreme heat events in complex urban settings. This approach provided a pathway for enhancing urban resilience planning and sustainability in the face of intensifying climate extremes. The key findings of this study are as follows.

First, most areas on campus experienced very high levels of heat stress under extreme conditions, with paved surfaces exhibiting the highest thermal burden. In contrast, tree canopies and water bodies provided substantial cooling benefits, highlighting the need for sustainable design strategies that enhance thermal comfort while fostering heat resilience.

Second, the ST-ViT model showed its ability to integrate multimodal data and achieve fine-scale spatiotemporal predictions of the UTCI. It outperformed baseline models in accuracy, achieving an MAE of 1.77°C and a computational efficiency 16 times greater than physics-based models (excluding preprocessing time). These capabilities enable real-time predictions and seamless integration into digital twin platforms, offering a scalable and efficient tool for climate-responsive planning.

Third, the digital twin platform further equips stakeholders with actionable insights for proactive heat stress planning and management. Its features, such as thermal comfort predictions and route recommendations, enhance safety and mobility under extreme heat conditions, providing practical solutions to mitigate risks and improve campus heat resilience.

## Acknowledgments

We greatly appreciate the helpful comments and suggestions from the editor and anonymous reviewers. The research was supported by National Science Foundation (NSF) under grant CMMI-2430700 and CNS-2401860, NASA under 80NSSC22KM0052, and Texas A&M University

Internal Funding. The funders had no role in the study design, data collection, analysis, or preparation of this article.

**Integrating Spatiotemporal Vision Transformer into Digital Twins for High-Resolution Heat Stress Forecasting in Campus Environments**

Supplemental Material

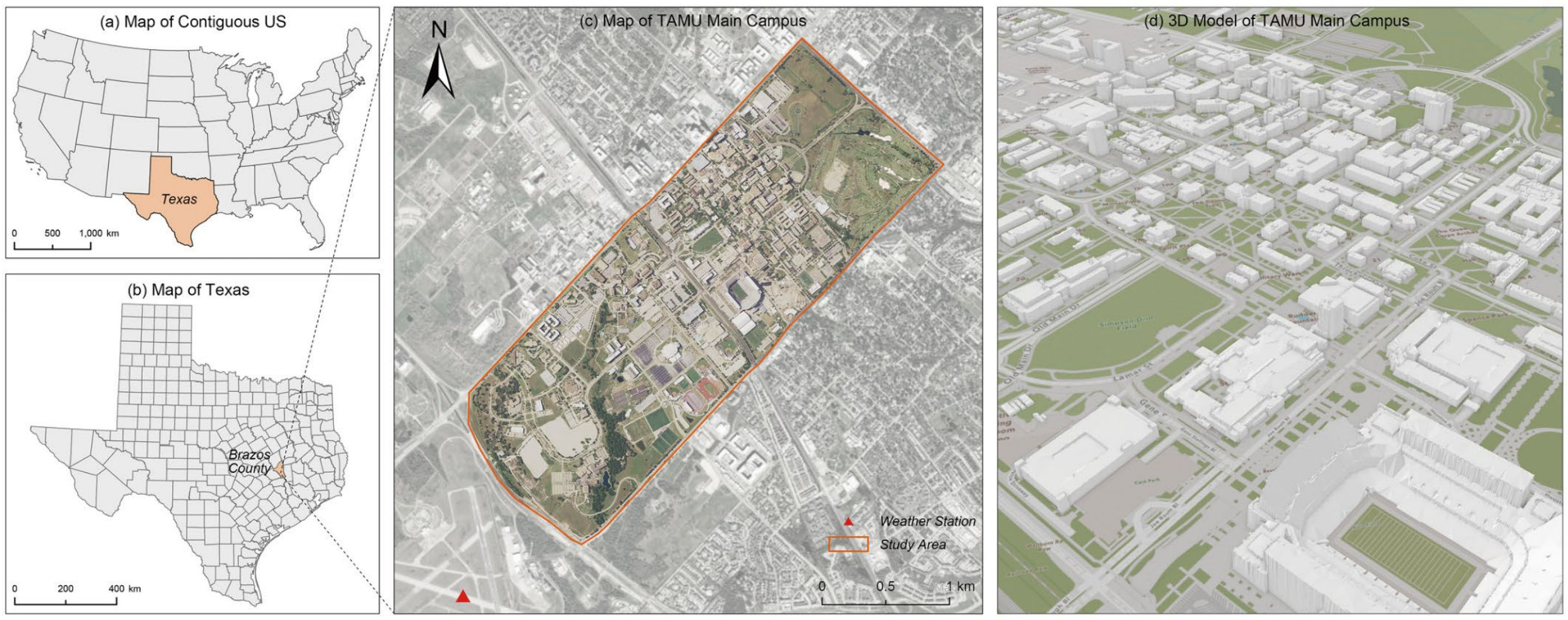

Fig. S1. Study area. (a) Texas is located in the southern US. (b) TAMU, located in Brazos County, is in the eastern part of Texas. (c) Map of TAMU main campus. (d) 3D model of TAMU main campus.

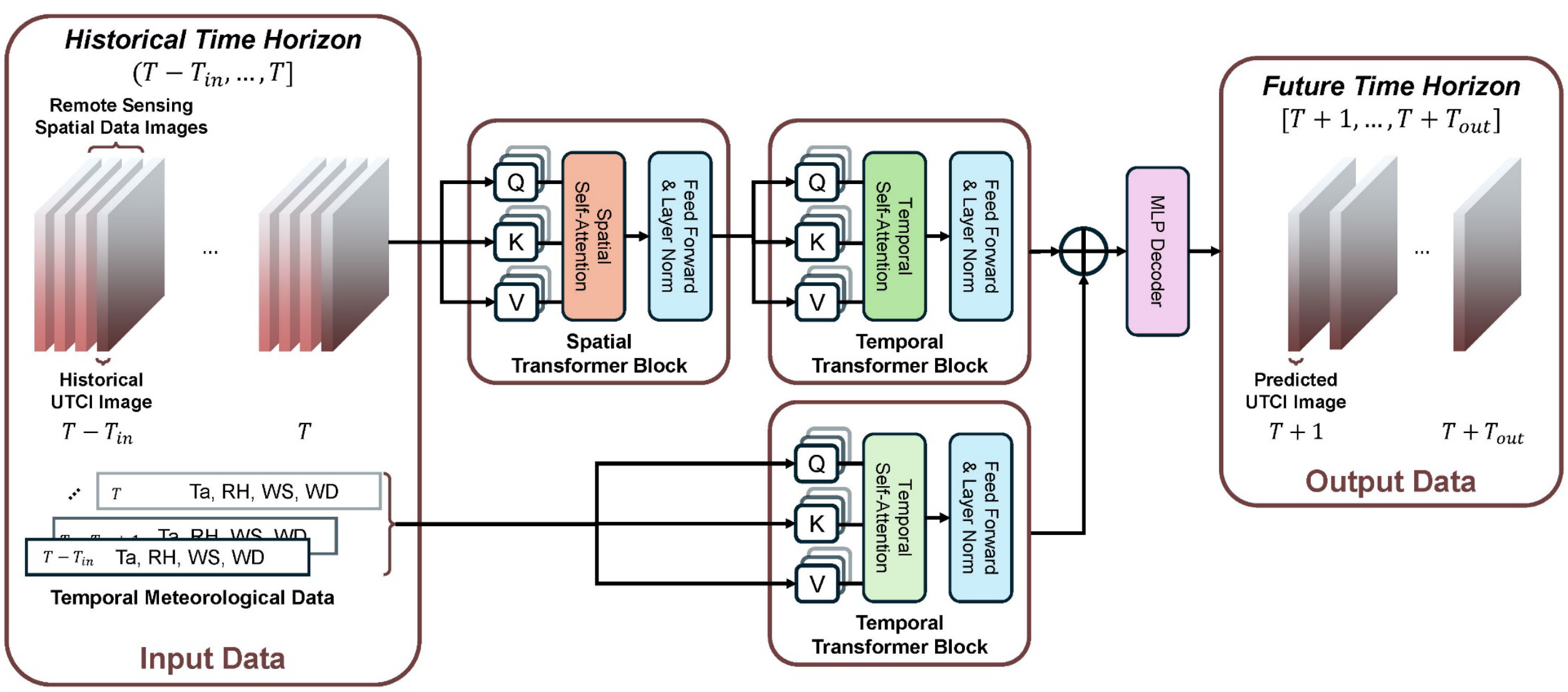

Fig. S2. Overall diagram of the ST-ViT model. The architecture employs three parallel attention streams to independently capture different dependencies in the input data: (1) The Spatial Multi-Head Attention applies attention across the spatial dimensions ($H \times W$) for each time step ($T_{\text{in}}$). This mechanism captures local and global spatial relationships, such as interactions between buildings and trees within a single time step. The query ($Q$), key ($K$), and value ($V$) matrices are defined as: $Q, K, V \in \mathbb{R}^{(H \cdot W) \times d_{\text{attn}}}$, where $H \times W$ represents the flattened spatial grid, and $d_{\text{attn}}$ is the attention feature dimension. (2) The Temporal Multi-Head Attention focuses on temporal dependencies across time steps ($T_{\mathrm{in}}$) for each spatial location. This mechanism captures temporal trends and long-term dependencies of human heat stress conditions. For this stream, the $Q$, $K$, and $V$ matrices are defined as: $Q, K, V \in \mathbb{R}^{T_{\mathrm{in}} \times d_{\mathrm{attn}}}$, where $T_{\mathrm{in}}$ is the number of historical time steps, and $d_{\mathrm{attn}}$ is the attention feature dimension. (3) Another Temporal Multi-Head Attention independently processes temporal meteorological features, modeling inter-variable correlations (e.g., air temperature and wind speed) over time. The attention operates similarly to the Temporal Multi-Head Attention stream but exclusively focuses on meteorological data. Each attention layer employs multi-head self-attention mechanisms to capture complex dependencies, while feed-forward networks, layer normalization, and residual connections ensure stable training and effective propagation of features. After parallel processing, the spatial-temporal features and temporal features are fused using an additive mechanism. Temporal features are expanded to match the spatial resolution before being combined with the spatial-temporal features, resulting in a unified representation. The final output is produced through a linear projection layer, which transforms the fused hidden representation into the desired output channels while preserving the spatial and temporal resolution of the task.

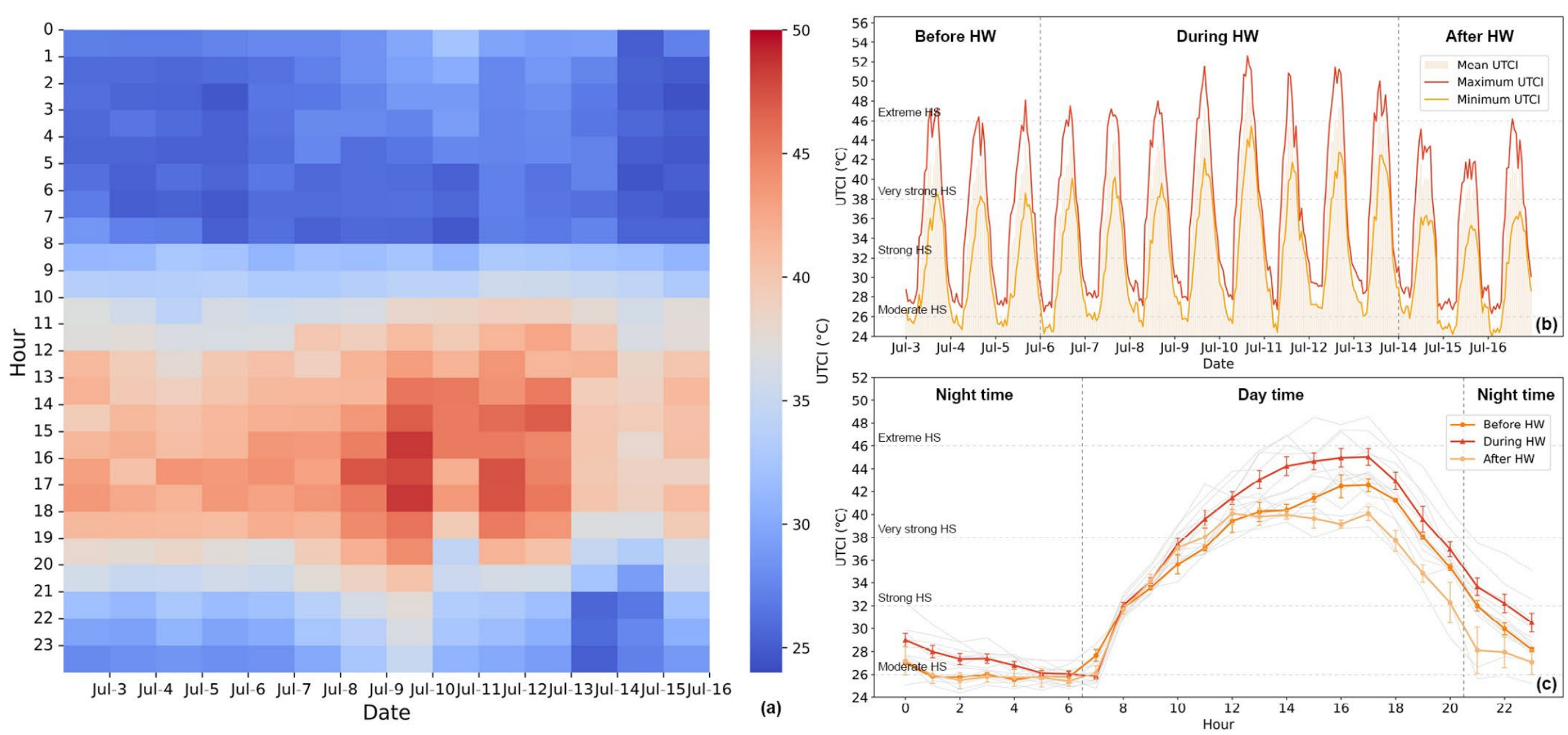

Fig. S3. Temporal variation in UTCI during different heatwave stages based on SOLWEIG simulations. (a) Heat map of hourly mean UTCI values from July 3 to 16, 2022. (b) Daily trends of minimum, mean, and maximum UTCI. (c) Hourly mean UTCI variation averaged by heatwave stage.

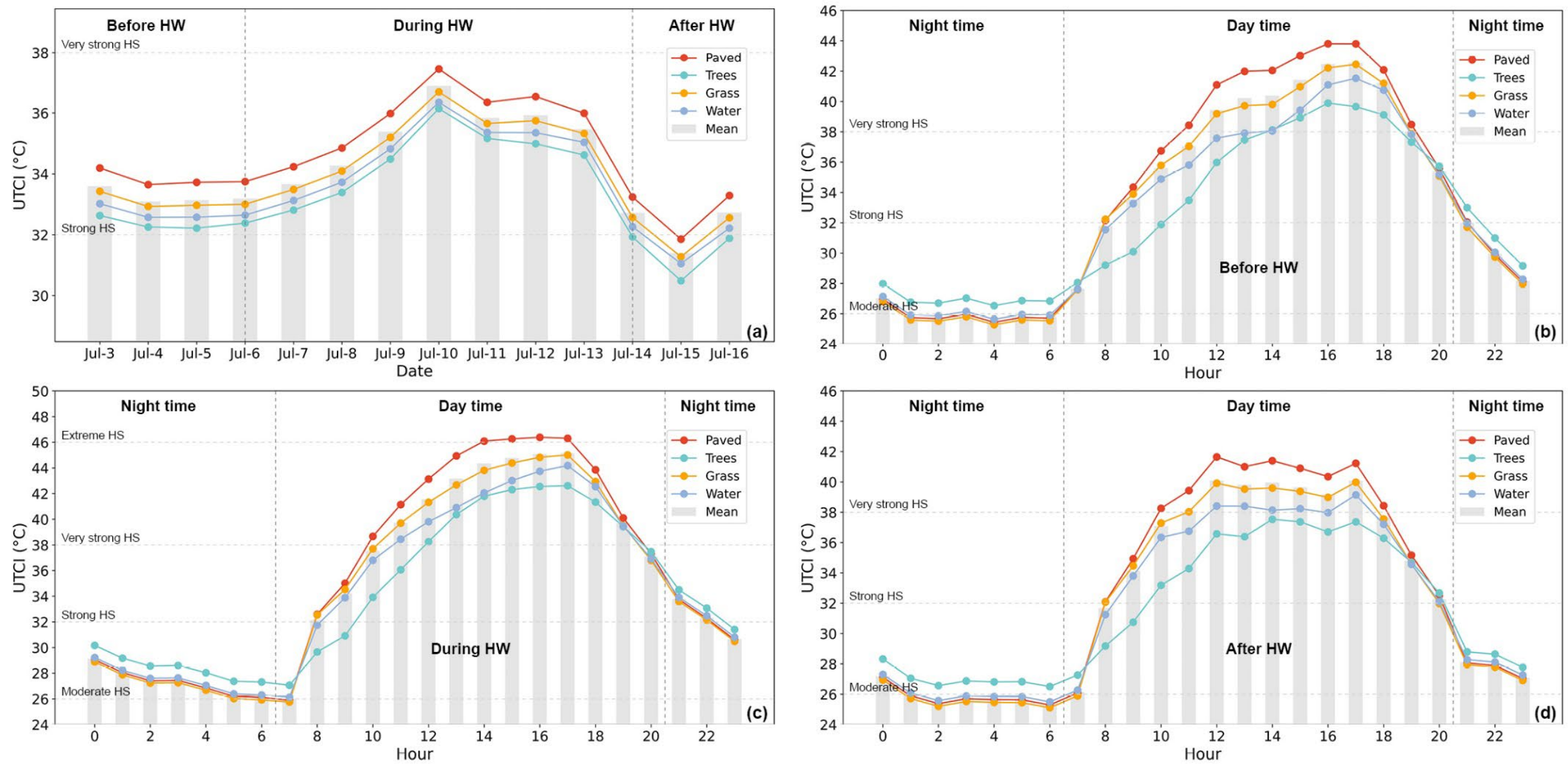

Fig. S4. UTCI variation across land cover types during different heatwave stages based on SOLWEIG simulations. (a) Daily UTCI from July 3 to 16, 2022; (b–d) Diurnal UTCI patterns before, during, and after the heatwave.

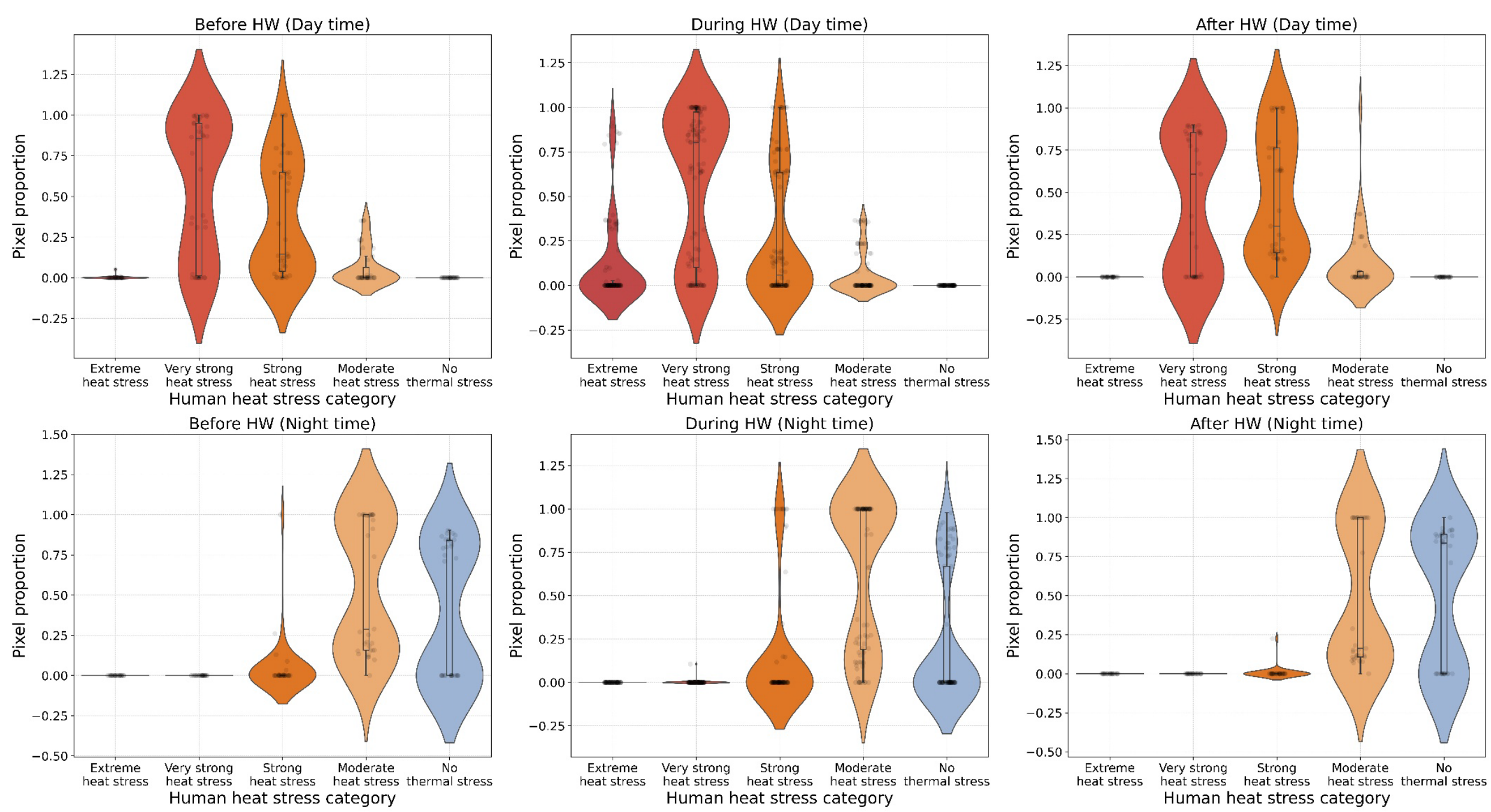

Fig. S5. Distribution of pixel proportions across UTCI thermal stress categories based on SOLWEIG simulations. Each dot represents the value at one hourly timestep.

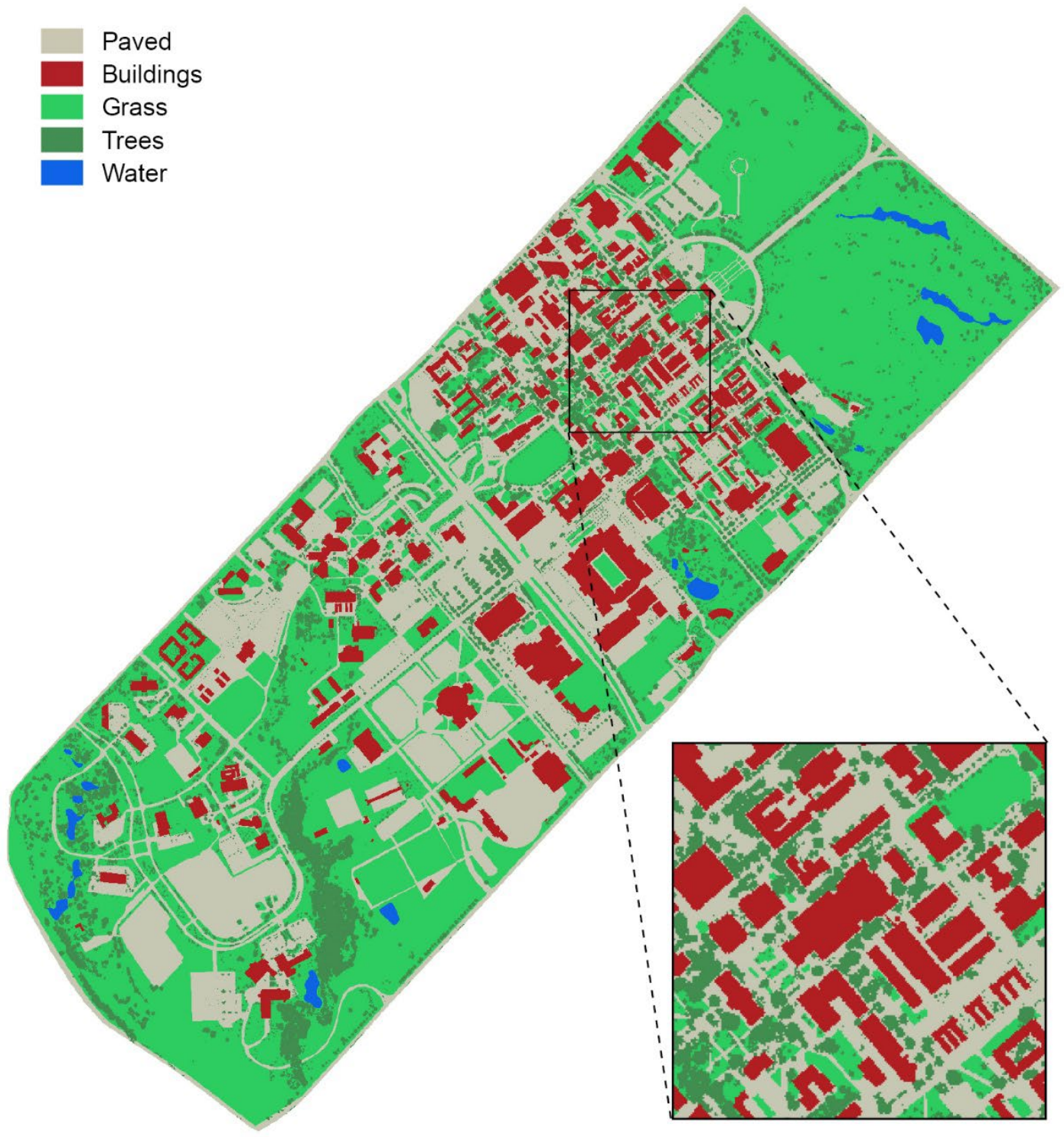

Fig. S6. Landcover classes of the study area.

Table S1. Input data characteristics and roles in the ST-ViT model and digital twin.

| Time horizon | Data category | Feature | Resolution | Format | Used in the model for |
|---|---|---|---|---|---|
| Historical | Spatial | Digital elevation model | Spatial: 1 m; Temporal: static | Raster | Spatial encoder |
| | | Building height model | Spatial: 1 m; Temporal: static | Raster | Spatial encoder |
| | | Tree height model | Spatial: 1 m; Temporal: static | Raster | Spatial encoder |
| | | Land cover | Spatial: 1 m; Temporal: static | Raster | Spatial encoder |
| | Meteorological | Air temperature | Spatial: campus-wide; Temporal: hourly | Tabular | Temporal encoder |
| | | Relative humidity | Spatial: campus-wide; Temporal: hourly | Tabular | Temporal encoder |
| | | Wind speed | Spatial: campus-wide; Temporal: hourly | Tabular | Temporal encoder |
| | | Wind direction | Spatial: campus-wide; Temporal: hourly | Tabular | Temporal encoder |
| | | Global horizontal irradiance | Spatial: campus-wide; Temporal: hourly | Tabular | Temporal encoder |
| | | Direct normal irradiance | Spatial: campus-wide; Temporal: hourly | Tabular | Temporal encoder |
| | | Diffuse horizontal irradiance | Spatial: campus-wide; Temporal: hourly | Tabular | Temporal encoder |
| | Spatiotemporal | UTCI maps | Spatial: 1 m; Temporal: hourly | Raster | Sequence input |
| Future | Spatiotemporal | UTCI maps | Spatial: 1 m; Temporal: hourly | Raster | Model output |

Note: The historical time horizon refers to the 24-hour input window per sample ($T_{in}$ = 24), while the future horizon corresponds to the predicted 24-hour UTCI sequence ($T_{out}$ = 24) .

Table S2. Deep learning model parameters.

| Model | Model parameters |
|---|---|
| CNN+LSTM | batch_size=16, kernel_size=3, hidden_channels=64, num_layer=2, ReLU, num_epochs=200, patience=10, min_delta=0.0005 |
| U-Net+LSTM | batch_size=16, kernel_size=3, hidden_channels=64, num_layer=2, ReLU, num_epochs=200, patience=10, min_delta=0.0005 |
| ST-ViT | batch_size=10, hidden_dim=12, num_heads=2, num_layers=1, ReLU, num_epochs=200, patience=10, min_delta=0.0005 |

Table S3. UTCI ranges and corresponding thermal stress categories (Source: Błażejczyk et al., 2013).

| UTCI range (◦C) | Thermal stress category |
|---|---|
| > 46 | Extreme heat stress |
| 38 to 46 | Very strong heat stress |
| 32 to 38 | Strong heat stress |
| 26 to 32 | Moderate heat stress |
| 9 to 26 | No thermal stress |
| 0 to 9 | Slight cold stress |
| -13 to 0 | Moderate cold stress |
| -27 to -13 | Strong cold stress |
| -40 to -27 | Very strong cold stress |
| < -40 | Extreme cold stress |

Table S4. Evaluation of the ST-ViT model at different time steps.

| Time step | 1 | 2 | 3 | 4 | 5 | 6 | 7 | 8 |
|---|---|---|---|---|---|---|---|---|
| RMSE (℃) | 2.301 | 2.225 | 2.170 | 2.163 | 2.180 | 2.202 | 2.250 | 2.305 |
| MAE (℃) | 1.850 | 1.801 | 1.784 | 1.770 | 1.793 | 1.812 | 1.857 | 1.892 |
| MAPE (%) | 6.052 | 5.903 | 5.833 | 5.811 | 5.854 | 5.910 | 6.035 | 6.106 |

Table S5. The overall accuracy of the deep learning models.

| Model | RMSE (℃) | MAE (℃) | MAPE (%) |
|---|---|---|---|
| CNN+LSTM | 3.696 | 2.552 | 6.568 |
| U-Net+LSTM | 2.870 | 2.453 | 6.257 |
| ST-ViT | 2.163 | 1.770 | 5.811 |

Note S1. SOLWEIG model configuration and justification for high-resolution UTCI simulation.

We used the SOLWEIG module included in UMEP (Universal Multi-scale Environmental Predictor) version 4.0.4 (released December 8, 2023) for QGIS 3.x, developed and maintained by Lindberg et al (Lindberg et al., 2008). This plugin integrates a suite of urban climate tools and supports both pre-processing and post-processing operations, and has been widely used in previous thermal comfort research (Ding et al., 2024; Kianmehr et al., 2023).

In this study, the SOLWEIG module was employed to simulate hourly mean radiant temperature ($T_{mrt}$) at a 1-meter spatial resolution. We followed the procedure outlined in the official UMEP documentation. In addition to supplying spatial and meteorological data, we adopted the default parameter settings, which align with recommended values for outdoor thermal comfort modeling in typical urban environments (Lindberg et al., 2008; Bröde et al., 2012) and reflect the material properties of common campus surfaces. Specifically, the settings include: shortwave radiation absorption = 0.70; longwave radiation absorption = 0.95; body posture = standing; wall emissivity = 0.90; ground emissivity = 0.95; wall albedo = 0.20; and ground albedo = 0.15.

Although SOLWEIG supports UTCI computation only at discrete point locations, it is widely used in thermal comfort research where UTCI is derived by combining simulated $T_{mrt}$ with other meteorological variables. As $T_{mrt}$ is the most influential parameter in UTCI calculation (Li et al., 2024), we consider that using high-resolution $T_{mrt}$ offers the highest feasible spatial fidelity for UTCI estimation when wind speed, relative humidity, and air temperature remain spatially coarser. Following prior studies (Basu et al., 2024; Li et al., 2024; Chaowen & Fricker, 2021), we developed a custom Python script to compute continuous UTCI maps at each hourly time step using SOLWEIG-simulated $T_{mrt}$ (1-meter resolution) and spatially uniform meteorological data. We acknowledge that this approximation may introduce simplifications due to the lack of high-resolution inputs for wind speed, relative humidity, and air temperature. Future efforts could improve accuracy by incorporating finer-scale meteorological observations.

To the best of our knowledge, there are currently no public tools that directly generate high-resolution UTCI maps. Most applications estimate UTCI indirectly by assembling the required input variables. Several existing tools are capable of simulating $T_{mrt}$, including ENVI-met, SOLWEIG, RayMan, CitySim, and Ladybug-Grasshopper. Among these, SOLWEIG stands out for its accessibility, offering a favorable trade-off between spatial resolution, computational efficiency, and usability, which makes it suitable for campus-scale thermal stress mapping (Merville et al., 2023).

Nevertheless, the SOLWEIG model has certain limitations, as documented in the official UMEP documentation. For instance, SOLWEIG focuses on pedestrian-level radiation fluxes and is not designed to simulate radiation fluxes on building roofs.